\documentclass[11pt]{article}

\usepackage[final]{acl}

\usepackage{times}
\usepackage{latexsym}

\usepackage[T1]{fontenc}

\usepackage[utf8]{inputenc}
\usepackage{adjustbox}

\usepackage{microtype}

\usepackage{inconsolata}

\usepackage{graphicx}
\usepackage{booktabs}
\usepackage{multirow}
\usepackage{tabularx}
\usepackage{array}

\usepackage{hyperref}

\usepackage{url}
\usepackage{hyperref}
\usepackage{booktabs}
\usepackage{multirow}
\usepackage{amsmath}
\usepackage{amssymb}
\usepackage{bbding}
\usepackage{wrapfig}
\usepackage{enumitem}
\usepackage{tabularx}
\usepackage{makecell}
\usepackage{array}
\newcolumntype{C}[1]{>{\centering\arraybackslash}m{#1}}
\newcolumntype{L}[1]{>{\raggedright\arraybackslash}m{#1}}
\usepackage{xcolor}         
\definecolor{ao}{rgb}{0.0, 0.5, 0.0}
\definecolor{forestgreen}{RGB}{0, 150, 0}
\definecolor{lightgraybox}{RGB}{240,240,240}

\usepackage{inconsolata} 
\usepackage{enumitem}
\usepackage{ragged2e}

\usepackage[most]{tcolorbox}
\newtcolorbox{promptbox}[1]{
  enhanced,
  breakable,
  colback=white,
  colframe=black,
  boxrule=0.6pt,
  arc=1pt,
  boxsep=0pt,
  left=5pt,
  right=5pt,
  top=3pt,
  bottom=3pt,
  before skip=2pt,
  after skip=2pt,
  title={#1},
  coltitle=white,
  colbacktitle=black,
  fonttitle=\bfseries\rmfamily\small,
  toptitle=2pt,
  bottomtitle=2pt,
  boxed title style={
    colback=black,
    colframe=black,
    boxrule=0pt,
    sharp corners
  }
}
\title{Beyond Routing Weights: Faithful Response-Level Interpretation of Mixture-of-Experts Reward Models via Contribution Contrast}

\author{\textbf{Yifan Wang\textsuperscript{1},}
        \textbf{Jinyi Mu\textsuperscript{2},}
        \textbf{Mayank Jobanputra\textsuperscript{1},}
        \textbf{Yu Wang\textsuperscript{3},}
        \\
        \textbf{Soyoung Oh\textsuperscript{1}, }
        \textbf{Isabel Valera\textsuperscript{1,4}, }
        \textbf{Vera Demberg\textsuperscript{1,5}} 
        \vspace{0.6em}
        \\
        \textsuperscript{1} Saarland University  \space \textsuperscript{2} University of California San Diego \space
        \textsuperscript{3} Utrecht University \\
        \textsuperscript{4} Max Planck Institute for Software Systems \space
        \textsuperscript{5} Max Planck Institute for Informatics \vspace{0.4em} \\
        \texttt{yifwang@lst.uni-saarland.de}
}

\begin{document}
\maketitle
\begin{abstract}
Reward models are central to learning from human preferences, yet identifying what drives their predictions remains challenging. 
Recent sparse Mixture-of-Experts (MoE) reward models seek to improve interpretability by routing prompts to specialized experts and characterizing experts through examples with high routing weights.
However, routing weights only reveal which prompts an expert \textit{receives}, not how it \textit{judges} responses, providing only a partial account of expert behavior. 
We therefore propose \textbf{Co}ntribution-\textbf{Co}ntrast (\textbf{CoCo}) response-level interpretation, which faithfully characterizes experts' roles using chosen--rejected response pairs with the largest contribution contrasts, jointly capturing routing and preference behavior. 
Across automatic and human evaluations, CoCo yields more coherent and faithful interpretations than router-based, score-based, and sparse autoencoder-based alternatives.
Overall, we provide a controlled study of how different example-selection signals affect the interpretability of MoE reward models.

\end{abstract}

\section{Introduction}

Reward models play a central role in reinforcement learning from human feedback
(RLHF) and in aligning large language models (LLMs) with human preferences~\citep{rlhf-bai}. 
However, understanding which properties drive their preference predictions remains
challenging~\citep{luo-etal-2025-rethinking}.
Recent work therefore seeks to discover interpretable preference patterns
directly from standard preference data, without relying on costly fine-grained annotations~\citep{movva2026whats}.

Sparse mixture-of-experts (MoE) reward models offer a promising approach by using a prompt-conditioned router to assign inputs to specialized response-scoring experts~\citep{wang2026sparse}. 
With sparsity and diversity regularization, these models learn coherent routing patterns and experts specialized in distinct domains. 
However, the choice of signal for ranking and selecting examples to interpret MoE experts remains underexplored.
\citet{wang2026sparse} interpret experts using examples with the highest routing weights, but these weights reveal only which prompts an expert \textit{receives}, not how it \textit{judges} responses. 
Such interpretations thus provide only an incomplete account of experts' roles in MoE, and primarily capture the prompt-level domains in which each expert specializes, rather than the response-level preference dimensions that directly explain the model’s decisions.
We discuss the related work in more detail in Appendix~\ref{appendix:related_work}.

To address these limitations, we propose \textbf{CoCo}, a \textbf{Co}ntribution \textbf{Co}ntrast-based response-level interpretation method for MoE reward models. 
CoCo uses contribution contrast as the interpretation signal and characterizes experts using response pairs with a large product of the routing weight and the absolute expert score difference, thereby capturing the interaction between where an expert is used and how it distinguishes responses. 
We further adapt existing interpretability regularizers to encourage sparse and diverse contribution contrast patterns.
Across automatic and human evaluations on two datasets, CoCo
produces more coherent, faithful, and specialized interpretations than
router-based, score-based, and sparse autoencoder (SAE) alternatives while
maintaining competitive reward modeling accuracy. Controlled comparisons
within the same MoE and across signal-specific training objectives further
show that CoCo provides a more informative account of expert
behavior than routing weights or expert score differences alone.

\begin{table}[t]
\centering
\small
\begin{tabularx}{\columnwidth}{l X}
\toprule
\textbf{Method} & \textbf{Interpretation} \\
\midrule
SAE-based &
Captures response-level features, but their weights are typically fixed, making the method less suitable for modeling flexible input-dependent preferences. \\
\midrule
Router-based &
Captures prompt-level domains or contexts in which each expert is used, but not how it distinguishes between responses. \\
\midrule
Score-based &
Captures response-level preference dimensions, but may emphasize examples where the expert has little influence because of low routing weight. \\
\midrule
CoCo &
Captures response-level preference dimensions that are faithful to model decisions by accounting for the interaction between routing weights and expert scores.\\
\bottomrule
\end{tabularx}
\caption{Conceptual comparison of different interpretability signals used for interpreting reward models.} 
\label{tab:interpretation_summary}
\vspace{-1.0em}
\end{table}

\section{Interpreting MoE Reward Models via CoCo}
\label{sec:method}

We first introduce the MoE reward model architecture, the contribution contrast-based interpretability regularizers, and the procedure for generating CoCo interpretations. 
We also define two alternative MoE interpretation methods using router-based and score-based signals as comparison baselines.

\paragraph{MoE Reward Model}


To capture heterogeneous and interpretable preference patterns, prior work implements reward models using the MoE architecture~\citep{wang-etal-2024-interpretable, shen-etal-2025-micro}. 
Given a preference pair $(x,y_w,y_l)$, an MoE reward model consists of a prompt-conditioned router
$\boldsymbol{\pi}_\phi(x)\in\Delta^{K-1}$ and $K$ reward experts
$\{r_{\theta_k}\}_{k=1}^{K}$. 
The router produces a softmax distribution over experts, where $\pi_{\phi,k}(x)$ denotes the weight assigned to expert $k$. 
The final reward is represented as the weighted average of expert scores:
\begin{equation}
r(x,y)
=
\sum_{k=1}^{K}
\pi_{\phi,k}(x)\,
r_{\theta_k}(x,y).
\end{equation}
Conventionally, the MoE reward model is trained on a preference dataset $\mathcal{D}=\{(x,y_w,y_l)\}$ using the Bradley--Terry
objective.

\paragraph{CoCo-Based Interpretability Regularization}

\citet{wang2026sparse} introduce three interpretability regularizers for learning disentangled and specialized experts: local sparsity
$\mathcal{L}_{\mathrm{ls}}$, global balance
$\mathcal{L}_{\mathrm{gb}}$, and expert diversity
$\mathcal{L}_{\mathrm{div}}$. 
These regularizers respectively promote sparse routing weights, balanced expert utilization, and diverse expert behavior.
Since they use routing weights as the interpretability signal, these regularizers primarily operate on $\pi_\phi(x)$. 
CoCo instead interprets experts through their contribution contrasts in the final rewards and thus applies the regularizers to pairwise contribution contrasts.
For expert $k$, we define its signed contribution contrast and magnitude as:
\begin{equation}
\begin{aligned}
\Delta c_k(x,y_w,y_l)
=&\pi_{\phi,k}(x)
\left[r_{\theta_k}(x,y_w)-r_{\theta_k}(x,y_l)\right],\\
c_k(x,y_w,y_l)
&=\left|\Delta c_k(x,y_w,y_l)\right|.
\end{aligned}
\end{equation}
The magnitude $c_k$ measures how strongly the expert distinguishes the
responses, while the sign of $\Delta c_k$ indicates which response it
supports.

Let $\mathbf{c}(x,y_w,y_l)=[c_1,\ldots,c_K]$ denote the contribution contrast profile of a preference pair. We apply the sparsity, balance, and diversity regularizers to these profiles rather than to routing weights, encouraging interpretable and distinct contribution patterns (formal definitions are provided in Appendix~\ref{appendix:regularizers}).
This lightweight adaptation leaves the MoE architecture unchanged, while aligning the interpretability constraints with the contrastive contribution signal used by CoCo.
An ablation study in Section~\ref{sec:coco_training} analyzes the effect of the CoCo-based interpretability regularization.

\paragraph{CoCo Interpretation Extraction}

Following \citet{movva2026whats}, who show that preferences are better characterized through contrastive examples, we interpret each MoE expert using chosen--rejected response pairs. 
For expert $k$, we select the 10 validation examples with the largest contribution contrast magnitudes $c_k(x,y_w,y_l)$ and prompt an LLM to summarize their shared contrastive pattern. 
We provide the input prompt and both responses, and indicate which response receives the larger contribution from the target expert. 
The resulting interpretation may capture a shared context, a consistent preference dimension, or both.

\paragraph{Alternative Interpretation Signals}

While CoCo captures the interaction between routing and expert behavior, we define two alternative interpretation signals as comparison baselines: router-based and score-based interpretation. When interpreting expert $k$, router-based interpretation ranks examples by the routing weight $\pi_{\phi,k}(x)$,\footnote{The router-based baseline is identical to the interpretation method used in \citet{wang2026sparse}.} whereas score-based interpretation ranks them by the absolute expert score difference $|r_{\theta_k}(x,y_w)-r_{\theta_k}(x,y_l)|$.

\begin{table*}[ht]
    \centering
    \resizebox{\textwidth}{!}{
    \begin{tabular}{llccccccc}
        \toprule
        \multirow{2}{*}{Dataset} &
        \multirow{2}{*}{Method} &
        \multirow{2}{*}{RM Acc.} &
        \multicolumn{2}{c}{Interpretation Quality} &
        \multicolumn{2}{c}{Decision Faithfulness} &
        \multicolumn{2}{c}{Expert Specialization} \\
        
        \cmidrule(lr){4-5}
        \cmidrule(lr){6-7}
        \cmidrule(lr){8-9}
        
        & & &
        Fidelity &
        Redundancy ($\downarrow$) &
        EM Agree. &
        Removal Flip &
        Expert Acc. &
        Expert Adv. \\
        \midrule

                \multirow{4}{*}{700K}
        & SARM
        & 75.72 & 0.063 & 0.654
        & 64.44 & 0.66 & 52.31 & 0.95 \\
        
        & WIMHF
        & 60.43 & 0.202 & \textbf{0.516}
        & 59.65 & \textbf{7.19} & 54.25 & -7.63 \\
        
        & SMoE
        & \textbf{84.43} & 0.264 & 0.593
        & 87.92 & 0.64 & 78.28 & \textbf{17.84} \\
        
        \cmidrule(lr){2-9}
        
        & CoCo 
        & 83.55 & \textbf{0.357} & 0.647
        & \textbf{97.40} & 1.29
        & \textbf{94.40} & 1.68 \\
        
        \midrule

        \multirow{4}{*}{Reddit}
        & SARM
        & 62.73 & 0.162 & 0.659 & 68.26 & 1.40 & 56.56 & 0.69 \\
        
        & WIMHF
        & 65.47 & 0.232 & \textbf{0.516}
        & 62.62 & \textbf{8.94} & 55.97 & 0.52 \\
        
        & SMoE
        & \textbf{70.40} & 0.344 & 0.815
        & 61.61 & 2.18 & 52.21 & 0.07 \\
        
        \cmidrule(lr){2-9}
        
        & CoCo
        & 68.64 & \textbf{0.421} & 0.669
        & \textbf{92.91} & 7.23
        & \textbf{85.09} & \textbf{3.24} \\
        
        \bottomrule

    \end{tabular}}
    \caption{Task performance and interpretability results of different interpretable reward models on 700K and Reddit. Best scores are marked in \textbf{bold}. CoCo achieves the strongest overall interpretability across both datasets.
    Results are averaged over five random seeds for SMoE and CoCo. 
    For SARM and WIMHF, we report single-seed results reproduced from the original papers.}
    \label{tab:interpretation_eval}
    \vspace{-1.5em}
\end{table*}

\section{Experiments}

We show CoCo’s improved interpretability over baselines through automatic and human evaluations across two datasets. 
Appendix~\ref{appendix:examples} provides representative examples from different methods and analyzes how their content and style are affected by their respective interpretation signals.

\subsection{Experimental Setups}

\paragraph{Model} Following \citet{wang2026sparse}, we use
\href{https://huggingface.co/Ray2333/GRM-Llama3.2-3B-rewardmodel-ft}{GRM-Llama3.2-3B}
as a frozen backbone and train $K=20$ linear reward heads as experts.
The router is implemented as a one-hidden-layer MLP with 128 hidden units.
We fix the coefficients of
$\mathcal{L}_{\mathrm{ls}}$,
$\mathcal{L}_{\mathrm{gb}}$, and
$\mathcal{L}_{\mathrm{div}}$
to $0.5$, $1.0$, and $1.0$, respectively.
We evaluate CoCo on two binary preference datasets: \href{https://huggingface.co/datasets/hendrydong/preference_700K}{700K}~\citep{dong2024rlhf} and
\href{https://huggingface.co/datasets/stanfordnlp/SHP}{Reddit}~\citep{shp}.
The complete sets of CoCo interpretations on both datasets are provided in Appendix~\ref{appendix:examples}. 
Appendices~\ref{app:base-model} and \ref{app:ablation} present results with a different base model and an ablation study of the regularization weights.

\paragraph{Baselines} We compare CoCo with existing interpretable reward models and alternative interpretation signals for the MoE architecture.
For existing interpretable baselines, we include the sparse MoE reward model interpreted through routing weights
(SMoE; \citealp{wang2026sparse}).
We also include two SAE-based methods:
SARM~\citep{saerm} and WIMHF~\citep{movva2026whats}.
SARM first trains a sequence-level SAE encoder on unsupervised data and then fine-tunes a reward model on the target preference dataset.
WIMHF instead learns contrastive features directly from the embedding contrasts of chosen--rejected response pairs.
Both methods interpret their features by prompting an LLM to summarize highly activating examples.

To isolate the effect of the interpretation signal, we extract router-based, score-based, and CoCo interpretations from the same trained MoE model.
We also compare these interpretations on MoE variants trained with interpretability objectives targeted at routing weights, score contrasts, and contribution contrasts, respectively.
This allows us to distinguish the effect of post-hoc interpretation from that of the corresponding interpretability objective.
Table~\ref{tab:interpretation_summary} summarizes the conceptual differences between CoCo and baseline interpretation methods.
Implementation details are provided in Appendix~\ref{appendix:experimental_details}.
\subsection{Evaluation Metrics}
\label{sec:metrics}

We evaluate each method along three dimensions, with scores averaged across experts.
Note that while interpretation quality metrics directly measure interpretability, decision faithfulness and expert specialization assess how representative the selected examples are of the experts' roles, thus serving as indirect measures of interpretability.

\paragraph{Interpretation Quality}
We directly assess the quality of the LLM-generated interpretations.
\textbf{Fidelity} measures how well an interpretation generalizes to held-out examples, computed as the Spearman correlation between interpretation signal values and an LLM's judgments of whether each example follows the interpretation.
Higher correlation indicates stronger alignment between the expert signal and its interpretation.
\textbf{Redundancy} is the average pairwise cosine similarity between embeddings of expert interpretations. Lower redundancy indicates more distinct preference patterns.

\paragraph{Decision Faithfulness}
Since interpretations are derived from top-ranked examples, we evaluate whether these examples faithfully reflect the expert's role in the model's decisions.
\textbf{Expert--Model Agreement} (\textsc{EM Agree.}) measures the proportion of the top-ranked examples on which the target expert and the full MoE prefer the same response.
\textbf{Expert Removal Flip Rate} (\textsc{Removal Flip}) measures the proportion for which removing the target expert actually changes the MoE prediction, providing interventional evidence of the expert's influence.
High values on both metrics indicate that the selected examples capture behavior that is aligned with and consequential to the MoE's decisions.

\paragraph{Expert Specialization}
We likewise evaluate specialization on each expert's top-ranked examples.
\textbf{Expert Accuracy} (\textsc{Expert Acc.}) measures the target expert's preference accuracy on these examples.
\textbf{Relative Expert Advantage} (\textsc{Expert Adv.}) is the target expert's accuracy minus the mean accuracy of all other experts on the same set.
Together, these metrics capture both absolute competence and relative specialization of experts on their top-ranked examples.

To reduce sensitivity to the choice of selection threshold, we compute the decision faithfulness and expert specialization metrics at $p \in [0.01, 0.025, 0.05, 0.075, 0.1]$, where $p$ denotes the proportion of top-ranked examples, and report the normalized AUC.
We also report the reward model accuracy (\textsc{RM Acc.}) of each reward model.

\subsection{Comparison with Existing Baselines}
\label{sec:baseline_results}

Table~\ref{tab:interpretation_eval} shows that CoCo achieves the strongest overall interpretability among existing interpretable reward models across both datasets, obtaining the highest fidelity, expert--model agreement, and expert accuracy while retaining strong reward modeling accuracy.
On Reddit, it also achieves the largest relative expert advantage. 
While WIMHF yields lower redundancy and higher removal flip rates, it substantially underperforms on the remaining metrics.
SMoE attains a strong expert advantage on 700K, but trails CoCo in fidelity, decision faithfulness, and expert accuracy, and performs notably worse on Reddit.
Overall, CoCo yields the most coherent, faithful, and specialized expert interpretations while maintaining strong task performance, with no baseline outperforming it on more than two metrics.
A qualitative comparison in Appendix~\ref{appendix:examples} further shows that different methods capture distinct aspects of the experts' roles, reflecting the signals used for their interpretation.
\begin{figure}[h]
    \centering
    \includegraphics[width=\linewidth]{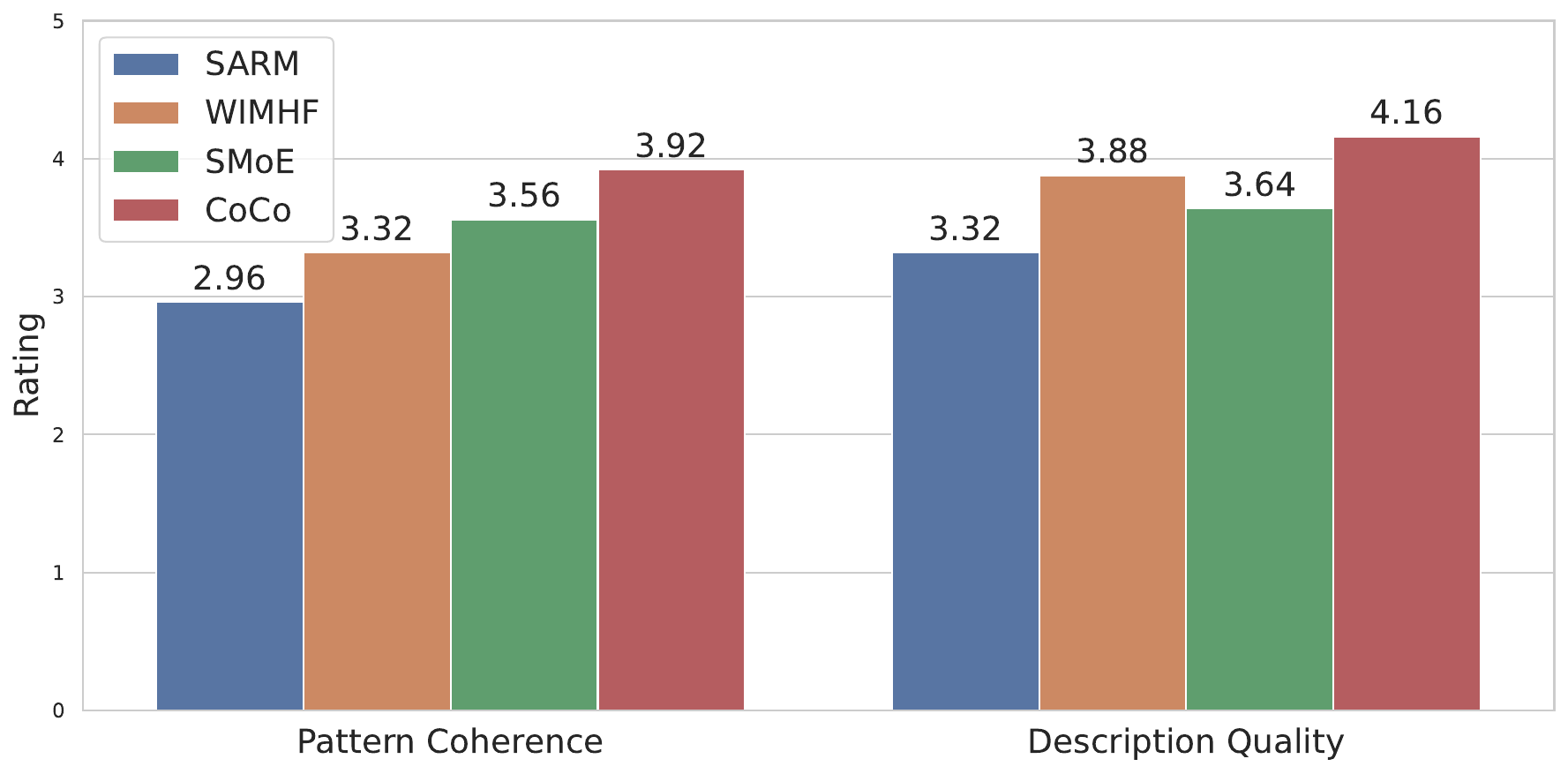}
    \caption{Human annotators rate CoCo highest in both pattern coherence and description quality.}
    \label{fig:human_study}
    \vspace{-1.0em}
\end{figure}
\paragraph{Human Evaluation} 
Following~\citet{wang2026sparse}, we sample five experts/features per method on Reddit and ask five annotators to evaluate their top ten examples for pattern coherence and description quality on 1--5 scales. 
Figure~\ref{fig:human_study} shows that CoCo produces the most coherent example sets for human annotators. 
Moreover, because its interpretations capture both shared contexts and response-level preference dimensions, they also receive the highest description quality ratings.

\subsection{Analysis of MoE Interpretation Signals}
\label{sec:comparison}

We compare CoCo with router- and score-based interpretation signals using their top-ranked Reddit examples. 
For a controlled comparison, we first apply all three signals to the same CoCo-regularized MoE. 
We then train separate MoE variants using signal-specific regularization (i.e., applying the same regularizers to different targets within the MoE architecture) and evaluate each interpretation signal on its corresponding model.

\vspace{-0.5em}
\begin{table}[ht]
    \centering
    \resizebox{\linewidth}{!}{
    \begin{tabular}{lcccc}
        \toprule
        
        \multirow{2}{*}{Signal} &
        \multicolumn{2}{c}{Decision Faithfulness} &
        \multicolumn{2}{c}{Expert Specialization} 
         \\
        
        \cmidrule(lr){2-3}
        \cmidrule(lr){4-5}

         &  
        EM Agree. &
        Removal Flip &
        Expert Acc. &
        Expert Adv. \\
        \midrule
        \multicolumn{5}{c}{MoE with CoCo Interpretability Regularization} \\
        \midrule
        Router-based & 77.86 & \textbf{8.29} & 64.24 & 1.21  \\
        Score-based & 85.59 & 7.51 & 64.92 & \textbf{6.58}  \\
        CoCo & \textbf{93.44} & 7.94 & \textbf{78.93} & 6.21  \\
        \midrule 
        \multicolumn{5}{c}{MoE with Signal-Specific Interpretability Regularization} \\
        \midrule
        Router-based & 57.61 & 1.28 & 51.15 & -0.39  \\
        Score-based & 62.20 & 2.57 & 58.13 & 5.51 \\ 
        CoCo & \textbf{93.44} & \textbf{7.94} & \textbf{78.93} & \textbf{6.21}  \\
  
        \bottomrule
    \end{tabular}}
    \caption{Reddit results of interpretations derived from different signals. Best scores are marked in \textbf{bold}.
    The CoCo signal yields top-ranked example sets that are more faithful to model decisions and exhibit stronger expert specialization than those selected by alternative interpretation signals.}
    \label{tab:comparison}
    \vspace{-0.5em}
\end{table}
Table~\ref{tab:comparison} shows that CoCo selects representative examples with the strongest overall decision faithfulness and expert specialization. 
On the same MoE, it achieves the highest expert--model agreement and expert accuracy, while remaining competitive in removal flip and relative expert advantage. 
Router- and score-based interpretations also remain substantially weaker when evaluated on MoE variants trained with their corresponding signal-specific regularization.
Overall, contribution contrast emerges as the most effective signal for selecting informative and representative example sets for MoE interpretation.

\subsection{Ablation of Interpretability Regularization Target}
\label{sec:coco_training}

In addition to comparing interpretation signals, we also analyze the effect of where the interpretability regularization is applied. 
Specifically, we consider MoE models trained with different signal-specific interpretability regularization.
We then use CoCo as the fixed signal for extracting interpretations.
This provides a complementary analysis to Section~\ref{sec:comparison}: rather than varying the interpretation signal, we keep it fixed and vary the target of interpretability regularization during training.

\begin{table}[ht]
    \centering
    \resizebox{\linewidth}{!}{
    \begin{tabular}{lcccc}
        \toprule
        
        \multirow{2}{*}{Regularization} &
        \multicolumn{2}{c}{Decision Faithfulness} &
        \multicolumn{2}{c}{Expert Specialization} 
         \\
        
        \cmidrule(lr){2-3}
        \cmidrule(lr){4-5}

         &  
        EM Agree. &
        Removal Flip &
        Expert Acc. &
        Expert Adv. \\
        \midrule
        
        Router-based & 64.49 & 0.16 & 59.91 & 4.43  \\
        Score-based & 61.71 & 0.12 & 58.18 & 4.80  \\
        CoCo & \textbf{93.44} & \textbf{7.94} & \textbf{78.93} & \textbf{6.21}  \\
  
        \bottomrule

    \end{tabular}}
    \caption{Reddit interpretability results of CoCo interpretations on MoE models trained with signal-specific interpretability regularization.
    Best scores are marked in \textbf{bold}.
    Applying interpretability regularization to contribution contrasts yields MoE reward models that produce more faithful and specialized CoCo interpretations.
    }
    \label{tab:coco_training}
\end{table}

Results in Table~\ref{tab:coco_training} shows that applying interpretability regularization to the contribution contrasts yields expert structures whose top-ranked examples more representatively reflect their respective experts' roles, resulting in more faithful interpretations.

\section{Conclusion}

We introduced \textbf{CoCo}, which interprets MoE reward model experts through contribution contrast, jointly capturing routing and expert preference behavior. 
Across automatic and human evaluations, CoCo produces more coherent, faithful, and specialized interpretations than router-based, score-based, and SAE-based alternatives while maintaining strong reward modeling accuracy. 
These findings support contribution contrast as a more informative basis for interpreting MoE reward models than routing weights or expert scores alone.

\section*{Limitations}

Our experimental evaluation has several limitations that should be acknowledged.
First, the quality of CoCo interpretations depends on the structure learned by the underlying MoE reward model. CoCo can expose how routing and expert scores jointly contribute to model decisions, but it cannot guarantee that the learned experts correspond to clean or semantically meaningful preference dimensions. As a data-centric approach, it may also fail to recover patterns that are absent, weakly represented, or insufficiently separable in the training data. Its effectiveness therefore remains sensitive to the dataset, model design, and training configuration.
Second, interpretability in our experiments is evaluated through indirect proxies, including fidelity, redundancy, decision faithfulness, expert specialization, and human judgments. Because the ground truth latent structure of preference data is unknown, these metrics cannot establish that an interpretation uniquely or completely characterizes an expert. Our results should therefore be understood as evidence that CoCo produces more behaviorally coherent and decision-aligned interpretations under the evaluated criteria, rather than as proof of ground truth disentanglement.

\section*{Ethical Considerations}

CoCo is intended as an auditing tool for understanding reward model behavior, not as evidence that the learned expert patterns represent desirable or universally shared human values. 
Since the method is data-centric, its interpretations may reflect biases, harmful content, or demographic imbalances present in the underlying preference datasets. 
In particular, Reddit-derived examples may contain sensitive, offensive, or personally revealing content. 
We therefore avoid interpreting learned patterns as normative judgments or inferring user identities and attributes from them.

\section*{Acknowledgements}
This work was funded by the DFG project GRK 2853 "Neuroexplicit Models of Language, Vision, and Action" (project number 471607914).

\bibliography{custom}

\appendix

\clearpage
\section{Related Work}
\label{appendix:related_work}

\paragraph{Data-Centric Preference Decomposition}
Modeling human values has become important in LLM alignment using reinforcement learning methods~\citep{chen2026reinforcementlearningoptionhedging, hu2026adaptiveqlbsrlop, rlhf-bai}.
Human values are inherently pluralistic and diverse, especially in subjective tasks, such as bias or preference judgments \citep{wang-demberg-2024-parameter, wang-demberg-2024-rsa, wang2025bcos, wang2026anatomysociodemographicspluralisticoffensiveness, ICLR2026_c8a08327, maxmin-rlhf-2024, zhang2025diverging}.
Most reward models learn a single global preference function and therefore struggle to capture heterogeneous human values~\citep{reward-model-survey-2025-ijcai, reward-model-survey-2025-emnlp}. Prior work addresses this by conditioning reward models on annotated attributes, user identities, or predefined groups~\citep{wang-etal-2024-arithmetic, quan-2024-dmoerm, wang-etal-2024-interpretable, zhou-etal-2024-beyond, wu2026reward, zhao2024group, choi-etal-2025-copl, liu-etal-2025-llms, balepur-etal-2025-whose, chen2025pal}, but such supervision is costly and restricts personalization to known preference dimensions. More recent methods instead infer latent preference structure directly from standard binary chosen--rejected data using latent variables, prototypes, representation decomposition, or MoE models~\citep{poddar-vae-preference-2024, chidambaram2026direct, test-time-alignment-2024, luo-etal-2025-rethinking, shen-etal-2025-micro}. However, their learned components are not always semantically interpretable, limiting transparent personalization.

\paragraph{Interpretable Reward Models}

To address the limitations above, recent work has explored aligning reward models with human-interpretable preference patterns, especially with sparsity constraints~\citep{zhan2026adaptivesignalresuscitationchannelwise, zhan2026relativerepairabilitycalibrationbaseddiagnostic,zhan2026sparsityallocationshapeslabelfree}.
One line of work uses sparse autoencoders (SAEs) to decompose reward model representations into sparse latent features, which can then be interpreted through highly activating examples or used for preference modeling~\citep{movva2026whats, sparserm, saerm}.
While these methods provide fine-grained representation-level explanations, they are typically post-hoc: sparse features are extracted from a separately trained SAE encoder and are not necessarily optimized to align with the target preference distribution.
This two-stage pipeline also introduces additional training and analysis costs.
Moreover, SAE-based reward models often use a fixed set of feature weights across contexts, which limits their flexibility in modeling context-dependent preference patterns.

A more recent direction builds interpretability directly into the reward model through sparse Mixture-of-Experts (SMoE) architectures~\citep{wang2026sparse}.
SMoE reward models learn specialized experts from standard binary preference data with sparsity and diversity constraints and have been shown to improve both interpretability and personalization performance.
However,~\citet{wang2026sparse} characterize SMoE experts through examples with high routing weights, so the resulting interpretations capture prompt-level topic or task clusters, such as math~\citep{jobanputra2025llmssubtractnumbers}, finance~\citep{dai-etal-2026-realfin, dai2026llmsceosbenchmarkingstrategic}, or emergency avoidance~\citep{xu2026doeswildfirepredictionmodel}, rather than response-level preference patterns.
Moreover, they provide only a partial account of expert behavior, describing which examples an expert \textit{receives} but not how it \textit{judges} them.
In contrast, CoCo interpretations capture the interaction between routing patterns and expert behavior, providing a more faithful and informative account of response-level preference dimensions.

\section{Complete CoCo Interpretations \& Qualitative Analysis}
\label{appendix:examples}

Tables~\ref{tab:700k_expert_interpretations} and~\ref{tab:reddit_expert_interpretations} present the complete CoCo interpretations for all 20 experts on 700K and Reddit, respectively.
CoCo can capture both shared input contexts and response-level preference dimensions, and therefore yields diverse and informative expert interpretations.

\begin{table*}[t]
\centering
\small
\renewcommand{\arraystretch}{1.08}
\begin{tabularx}{\textwidth}{c X}
\toprule
\textbf{Expert} & \textbf{Interpretation} \\
\midrule
0  & Math problems involving Python answered with brief steps and a minimal calculation. \\
1  & Instruction-heavy prompts answered with structured, step-by-step lists. \\
2  & Step-by-step reasoning rather than brief answers. \\
3  & Task-specific, step-by-step reasoning rather than generic exposition. \\
4  & Thorough, domain-specific answers with concrete details rather than brief or dismissive replies. \\
5  & Detailed, structured explanations rather than refusals or deflections. \\
6  & Competitive-programming tasks answered with a numbered plan followed by Python code. \\
7  & Lengthy, conversational elaborations rather than concise, direct answers. \\
8  & Tailored, concrete solutions with examples rather than vague or generic replies. \\
9  & Detailed instructions followed with specific, task-complete outputs rather than vague summaries. \\
10 & Generic, textbook-style exposition rather than direct answers to the prompt. \\
11 & Structured, detailed, and example-backed explanations. \\
12 & Exact adherence to output-format constraints. \\
13 & Math word problems answered with step-by-step calculations and a boxed final answer. \\
14 & Informational prompts answered with concrete, example-backed explanations rather than vague meta-responses or deflection. \\
15 & Exact adherence to output-format constraints. \\
16 & Broad, tangential exposition rather than direct answers or instruction following. \\
17 & Correction of false premises rather than accepting them. \\
18 & Concrete, step-by-step guidance rather than vague generalities. \\
19 & Exact adherence to output-format constraints. \\
\bottomrule
\end{tabularx}
\caption{CoCo interpretations of the 20 experts on 700K.}
\label{tab:700k_expert_interpretations}
\end{table*}

\begin{table*}[t]
\centering
\small
\renewcommand{\arraystretch}{1.08}
\begin{tabularx}{\textwidth}{c X}
\toprule
\textbf{Expert} & \textbf{Interpretation} \\
\midrule
0  & Science-fiction and comic-book lore questions answered with specific in-universe explanations. \\
1  & Concise, specific recommendations rather than lengthy lists or explanations. \\
2  & Cooking questions answered with references to specific authoritative sources and links. \\
3  & Practical, anecdote-backed explanations rather than abstract speculation. \\
4  & Energy and climate AMAs answered with skeptical questions about feasibility and trade-offs. \\
5  & Detailed, experience-based advice with concrete steps rather than brief replies. \\
6  & Fiction and fandom questions answered with witty, in-universe explanations. \\
7  & Personal, experience-based advice with specific reasoning rather than one-line suggestions. \\
8  & Blunt, sardonic humor rather than polite, measured replies. \\
9  & Specific, concrete answers rather than vague generalities. \\
10 & Concise, authoritative answers with actionable advice rather than personal anecdotes. \\
11 & Witty, humorous replies rather than detailed step-by-step advice. \\
12 & Practical advice questions answered with specific, experience-based details rather than generic replies. \\
13 & Concrete, actionable advice with specific examples or mechanisms rather than generic opinions. \\
14 & Legal advice questions answered with statute citations and actionable steps. \\
15 & Legal or HR dispute questions answered with assertive, actionable escalation advice. \\
16 & Advice questions answered with blunt, decisive guidance rather than hedged or meandering replies. \\
17 & Procedural and technical explanations that correct misconceptions rather than provide vague reassurance. \\
18 & Concise, opinionated tips rather than recipe links or lengthy lists. \\
19 & First-person, authoritative explanations with specific details rather than terse, impersonal replies. \\
\bottomrule
\end{tabularx}
\caption{CoCo interpretations of the 20 experts on Reddit.}
\label{tab:reddit_expert_interpretations}
\end{table*}

Table~\ref{tab:qualitative_interpretations} presents representative interpretations produced by different methods.
Their content and style vary substantially, reflecting the signals used for interpretation.
CoCo is the only contrastive method that jointly considers prompt context and expert preference behavior, and thus captures both domains of specialization and response-level preference dimensions.
In contrast, router-based interpretation and SMoE consider only highly ranked prompts, producing descriptions of the contexts assigned to each expert.
Both SAE-based methods recover response-level features, but SARM operates on individual responses, whereas WIMHF summarizes chosen--rejected pairs.
Consequently, WIMHF yields contrastive descriptions while SARM generally does not.
Score-based interpretation also captures response-level expert behavior, but ignores routing weights. 
Its top-ranked examples are therefore often repetitive, leading to highly redundant interpretations.

\begin{table*}[t]
\centering
\small
\renewcommand{\arraystretch}{1.08}
\begin{tabularx}{\textwidth}{
    >{\centering\arraybackslash}m{0.11\textwidth}
    X
}
\toprule
\textbf{Method} & \textbf{Representative Interpretations} \\
\midrule

\multirow{5}{*}{\textbf{CoCo}}
& Science-fiction and comic-book lore questions answered with specific in-universe explanations. \\
& Legal advice questions answered with specific statute citations and actionable steps. \\
& Cooking questions answered with references to authoritative sources and links. \\
& First-person, authoritative explanations with specific details rather than terse, impersonal replies. \\
& Blunt, sardonic humor rather than polite, measured replies. \\
\midrule

\multirow{5}{*}{\textbf{Router-based}}
& Asks for in-universe lore or continuity analysis of a popular fictional franchise. \\
& Seeks baking advice or ideas, including recipes, techniques, ingredient pairings, or uses for leftovers. \\
& Solicits community recommendations, tips, or personal stories. \\
& Long, first-person advice-seeking posts describing complex real-life situations with extensive background details. \\
& Seeks practical next-step advice for a real-world problem, often legal or medical. \\
\midrule

\multirow{5}{*}{\textbf{Score-based}}
& Detailed, example-driven explanations rather than brief assertions. \\
& Concrete, experience-backed specifics rather than generic commentary. \\
& First-person anecdotes and practical, experience-based tips rather than terse or theoretical replies. \\
& Advice questions answered with concrete, detailed, and actionable guidance. \\
& Terse, witty one-liners rather than detailed explanations. \\
\midrule

\multirow{5}{*}{\textbf{SMoE}}
& Asks a specific question seeking factual information, advice, or an opinion. \\
& Asks for legal advice or information about rights and laws. \\
& Asks a question or seeks advice in a Reddit-post format. \\
& Asks for troubleshooting or technical explanations of cooking or baking issues. \\
& Asks for advice or information about a specific profession or career. \\
\midrule

\multirow{5}{*}{\textbf{SARM}}
& Provides a direct, definitive answer to a factual or technical question. \\
& Provides a brief, direct, or conversational response. \\
& Provides an incorrect or factually hallucinated answer. \\
& Provides a step-by-step solution to a mathematics problem. \\
& Provides a brief, direct, or conversational response. \\
\midrule

\multirow{5}{*}{\textbf{WIMHF}}
& Offers advice and recommendations rather than asking questions, sharing anecdotes, or linking sources. \\
& Gives a terse, one-line reply without explanation or reasoning. \\
& Provides serious, information-oriented responses without jokes, insults, or generic encouragement. \\
& Includes specific recommendations or named examples. \\
& Provides detailed, multi-point advice with specific examples or resources. \\
\bottomrule
\end{tabularx}

\caption{Representative interpretations produced by CoCo, alternative MoE interpretation signals, and existing interpretable reward model baselines on Reddit.}
\label{tab:qualitative_interpretations}
\end{table*}

\section{Results with a Different Base Model}
\label{app:base-model}

To test whether our findings depend on the backbone, we replace
GRM-Llama3.2-3B with Skywork-Reward-V2-Qwen3-4B, a reward model from a
different model family. As before, the backbone is frozen, and we train the
router and $K{=}20$ linear experts on its hidden states on Reddit.

\begin{table*}[t]
\centering
\footnotesize
\setlength{\tabcolsep}{4pt}
\resizebox{\textwidth}{!}{%
\begin{tabular}{lccccccc}
\toprule

        \multirow{2}{*}{Method} &
        \multirow{2}{*}{RM Acc.} &
        \multicolumn{2}{c}{Interpretation Quality} &
        \multicolumn{2}{c}{Decision Faithfulness} &
        \multicolumn{2}{c}{Expert Specialization} \\
                
        \cmidrule(lr){3-4}
        \cmidrule(lr){5-6}
        \cmidrule(lr){7-8}
        
        & & 
        Fidelity &
        Redundancy ($\downarrow$) &
        EM Agree. &
        Removal Flip &
        Expert Acc. &
        Expert Adv. \\

\midrule
SMoE                    & 71.21 & 0.205 & 0.623 & 69.65 & \textbf{9.70} & 59.76 & 2.49  \\  
CoCo                    & \textbf{71.84} & \textbf{0.344} & \textbf{0.601} & \textbf{94.15} & 7.86 & \textbf{77.71} & \textbf{4.98} \\  
\bottomrule
\end{tabular}}
\caption{Results on Reddit with Skywork-Reward-V2-Qwen3-4B as the frozen backbone. Best values are in bold.}
\label{tab:base-model}
\end{table*}

Table~\ref{tab:base-model} shows that our main findings generalize to the new backbone. CoCo again outperforms SMoE on almost all metrics. Although SMoE achieves better Removal Flip, it performs substantially worse on EM Agree., suggesting weaker overall decision faithfulness. These results indicate that CoCo's interpretability advantages are not specific to the original backbone.

\section{Ablation of Expert Count and Regularization Weights}
\label{app:ablation}

We ablate the number of experts $K$ and the regularization weights
$(\lambda_{ls}, \lambda_{gb}, \lambda_{div})$ on Reddit. 
CoCo models are interpreted with the CoCo signal and SMoE with its routing weights.

\begin{table*}[h]
\centering
\footnotesize
\setlength{\tabcolsep}{6pt}
\begin{tabular}{llccccc}
\toprule
\multirow{2}{*}{Method} & \multirow{2}{*}{$K$} & \multirow{2}{*}{RM Acc.} & \multicolumn{2}{c}{Decision Faithfulness} & \multicolumn{2}{c}{Expert Specialization}  \\
\cmidrule(lr){4-5} \cmidrule(lr){6-7}
 & &  & EM Agree. & Removal Flip & Expert Acc. & Expert Adv. \\
\midrule
\multirow{5}{*}{SMoE}
& 5  & 68.16 & 85.09 & 26.71 & 63.53 & 5.06   \\  
& 10 & 71.10 & 69.69 & 11.14 & 58.18 & 2.84    \\  
& 20 & 72.71 & 57.61 & 1.28  & 51.15 & -0.36   \\  
& 30 & 70.42 & 59.13 & 1.01  & 48.78 & -0.16  \\  
& 50 & 70.77 & 58.38 & 2.17  & 51.49 & 0.06   \\  
\midrule
\multirow{5}{*}{CoCo}
& 5  & 70.03 & 99.70 & 6.63 & 86.92 & 1.85  \\  
& 10 & 69.42 & 98.06 & 0.06 & 91.74 & 0.55  \\  
& 20 & 70.74 & 93.44 & 7.94  & 78.93 & 6.21  \\  
& 30 & 70.91 & 93.46 & 0.50 & 86.57 & 4.76   \\  
& 50 & 71.57 & 86.11 & 3.09 & 74.39 & 6.69   \\  
\bottomrule
\end{tabular}
\caption{Effect of the number of experts $K$ on Reddit.}
\label{tab:ablation-k}
\end{table*}

\begin{table*}[h]
\centering
\footnotesize
\setlength{\tabcolsep}{4.5pt}
\begin{tabular}{cccccc}
\toprule
\multirow{2}{*}{$\lambda_{ls}$ / $\lambda_{gb}$ / $\lambda_{div}$} & \multirow{2}{*}{RM Acc.} & \multicolumn{2}{c}{Decision Faithfulness} & \multicolumn{2}{c}{Expert Specialization}  \\
\cmidrule(lr){3-4} \cmidrule(lr){5-6}
 &  & EM Agree. & Removal Flip & Expert Acc. & Expert Adv. \\
\midrule
0 / 0 / 0                          & 70.85 & 99.79 & 0.02  & 95.20 & 0.11    \\  
\midrule
\textbf{0.5 / 1.0 / 1.0}$^\dagger$ & 70.74 & 93.44 & 7.94  & 78.93 & 6.21    \\  
\midrule
0 / 1.0 / 1.0                      & 71.06 & 85.80 & 7.02  & 74.86 & 12.56   \\  
0.25 / 1.0 / 1.0                   & 71.42 & 86.89 & 8.21  & 74.84 & 10.14   \\  
1.0 / 1.0 / 1.0                    & 70.05 & 62.26 & 0.03  & 59.21 & 2.81   \\  
2.0 / 1.0 / 1.0                    & 70.51 & 67.53 & 0.00  & 63.61 & 1.90   \\  
\midrule
0.5 / 0 / 1.0                      & 70.68 & 57.78 & 0.00  & 57.11 & 4.18   \\  
0.5 / 0.5 / 1.0                    & 70.11 & 60.15 & 0.00  & 57.90 & 3.24   \\  
0.5 / 2.0 / 1.0                    & 70.80 & 87.81 & 7.07  & 75.33 & 9.69    \\  
\midrule
0.5 / 1.0 / 0                      & 70.92 & 94.32 & 2.40  & 85.01 & 5.13    \\  
0.5 / 1.0 / 0.5                    & 70.41 & 90.96 & 6.61  & 77.18 & 5.18    \\  
0.5 / 1.0 / 2.0                    & 70.38 & 89.19 & 10.03 & 75.24 & 6.79    \\  
\bottomrule
\end{tabular}
\caption{Effect of the regularization weights for CoCo on Reddit.
$\dagger$ denotes the setting used in the main experiments.}
\label{tab:ablation-lambda}
\end{table*}

\paragraph{Number of Experts}
Table~\ref{tab:ablation-k} shows that CoCo consistently outperforms SMoE in EM agreement and expert accuracy across all values of $K$. With more experts ($K>10$), CoCo generally achieves better removal flip and expert advantage as well. However, with fewer experts, CoCo underperforms on these two metrics, possibly because the limited number of experts prevents the formation of meaningful contribution structures, leading the model to rely primarily on a single dominant expert.

\paragraph{Regularization Weights}
Table~\ref{tab:ablation-lambda} varies one weight at a time.
Without regularization, all experts learn nearly identical reward directions: EM agreement is almost perfect (99.79), but the experts show little specialization (expert advantage: 0.11) or individual contribution to decisions (removal flip: 0.02). High agreement alone therefore does not imply interpretable experts, highlighting the importance of regularization.
Local sparsity improves the semantic coherence of experts but also encourages reliance on a single dominant expert, weakening decision faithfulness and expert specialization. Global balance counteracts this tendency by preventing expert collapse. Meanwhile, expert diversity encourages specialization across different domains, improving both removal flip and expert advantage.
Together, these results show that appropriately balancing the regularization terms helps prevent expert collapse and achieve better interpretability in MoE reward models.

\section{Experimental Details}
\label{appendix:experimental_details}

\paragraph{Datasets and Training} We adopt the same data pre-processing pipeline and training configuration as~\citet{wang2026sparse} for a controlled comparison.
\paragraph{LLM Usage for Interpretation Extraction}
Following~\citet{movva2026whats}, we use \texttt{gpt-5-low} to summarize top-ranked examples into expert interpretations and \texttt{gpt-5-mini-low} as the judge for computing fidelity. 
To measure redundancy, we encode all natural language interpretations using \href{https://huggingface.co/Qwen/Qwen3-Embedding-8B}{Qwen3-Embedding-8B} and compute their pairwise cosine similarities.

\paragraph{Baselines}

We train SARM and SMoE on our data splits following their official implementations. 
WIMHF is trained on its own pre-defined splits of the Reddit and 700K datasets and learns 32 features, compared with our 20 experts. 
We therefore evaluate WIMHF on its corresponding test splits and report interpretability results averaged over all 32 features.

\paragraph{LLM Instructions}
\label{appendix:prompts}

Figures~\ref{fig:prompt_describe_experts} and~\ref{fig:prompt_match_description} present the prompts used to generate LLM-based interpretations from top-ranked examples and to judge whether held-out examples match the resulting interpretations.

\begin{figure*}[p]
\centering
\begin{promptbox}{Prompt: Describing experts from top-ranked examples}
{\ttfamily
\fontsize{6.2}{6.9}\selectfont
\setlength{\parindent}{0pt}
\setlength{\parskip}{1pt}
\setlist[itemize]{
    leftmargin=1.5em,
    topsep=0pt,
    itemsep=0pt,
    parsep=0pt,
    partopsep=0pt
}

\textbf{Task Overview}

You are interpreting one component of a mixture-of-experts reward model.

The model contains a router and multiple reward experts. The router depends only
on the user prompt, whereas each expert score depends on the full
prompt--response sequence. An expert's contribution is its router weight
multiplied by its score. High contribution may reflect a prompt-level routing
pattern, a response-level scoring pattern, or both.

The examples below were selected because this component made a large
contribution to their final rewards. Identify the single most coherent pattern
associated with high contribution.

\textbf{Instructions}

Inspect the prompts and responses and identify the most specific pattern
consistently supported across the examples.

The pattern may describe:
\begin{itemize}[leftmargin=1.5em,nosep]
    \item a prompt-level pattern, such as topic, domain, task type, or user intent;
    \item a response-level property, such as style, reasoning, specificity,
          refusal behavior, or instruction following;
    \item a coherent combination of prompt type and response property.
\end{itemize}

Prefer one of the following forms:
\begin{itemize}[leftmargin=1.5em,nosep]
    \item ``\textless prompt or task type\textgreater''
    \item ``\textless response property\textgreater{} rather than
          \textless contrast\textgreater''
    \item ``\textless prompt or task type\textgreater{} answered with
          \textless response property\textgreater''
\end{itemize}

Examples:
\begin{itemize}[leftmargin=1.5em,nosep]
    \item ``science-fiction questions''
    \item ``gives concise recommendations rather than extended explanations''
    \item ``science-fiction questions answered with detailed in-universe reasoning''
    \item ``cooking questions answered with mechanistic explanations of ingredients''
    \item ``responds directly to sensitive requests rather than refusing''
    \item ``follows exact output-format constraints''
\end{itemize}

Use only properties recurring across multiple distinct examples. If only the
prompt topic is coherent, describe the routing pattern. If response behavior is
coherent across varied prompts, describe the response-level property. Do not
combine unrelated clusters.

\textbf{Additional Rules}

\begin{itemize}[leftmargin=1.5em,nosep]
    \item Be objective, concrete, and concise.
    \item Identify one pattern, not a list of possible properties.
    \item Do not treat duplicated prompts as independent evidence.
    \item Do not assume high contribution implies higher quality or correctness.
    \item Avoid vague labels unless made behaviorally specific.
    \item Avoid ``either,'' long lists, and unrelated clauses.
    \item Do not mention the expert, router, score, contribution, activation,
          or examples in the final answer.
\end{itemize}

\textbf{Examples}

----------------\\[-2pt]
\{examples\}\\[-2pt]
----------------

\textbf{Output}

Output exactly one concise description, beginning with ``-'' and enclosed in
quotation marks.

Your response is:-"
}
\end{promptbox}
\caption{Instruction used to summarize the common feature of top-ranked examples.}
\label{fig:prompt_describe_experts}
\end{figure*}

\begin{figure*}[t]
\centering
\begin{promptbox}{Prompt: Judging whether an example matches expert description}
\ttfamily\scriptsize

You are evaluating a response-level interpretation of a reward-model expert.

\medskip
\textbf{Interpretation}

\{description\}

\medskip
Both responses answer the same user prompt. Decide which response expresses the interpretation more strongly, taking the prompt into account.

\medskip
\textbf{Prompt}

\{prompt\}

\medskip
\textbf{Response A}

\{response\_a\}

\medskip
\textbf{Response B}

\{response\_b\}

\medskip
\textbf{Instructions}

Output exactly one of the following labels:
\begin{itemize}[leftmargin=2em, itemsep=2pt, topsep=2pt]
    \item \texttt{A}: Response A expresses the interpretation more strongly.
    \item \texttt{B}: Response B expresses the interpretation more strongly.
    \item \texttt{NOT\_RELEVANT}: The interpretation does not distinguish the responses, both express it equally, neither expresses it, or the evidence is unclear.
\end{itemize}

Consider general response quality only when it is part of the interpretation. Do not output any explanation or additional text.

\medskip
Answer:
\end{promptbox}
\caption{Instruction used to judge which response in a pair more strongly matches a given natural language expert interpretation.}
\label{fig:prompt_match_description}
\end{figure*}

\paragraph{Human Study}
We adopt the same human evaluation pipeline as in~\citet{wang2026sparse}.
Annotators were compensated at an hourly rate comparable to the local average hourly salary for researchers and informed about how their data would be used.

\section{CoCo-Based MoE Reward Model Training}
\label{appendix:regularizers}

Following \citet{shen-etal-2025-micro} and \citet{wang2026sparse}, we train the model by marginalizing the expert-level Bradley--Terry probabilities:
\begin{equation} \begin{aligned} \mathcal L_{\mathrm{RM}} = -\frac{1}{|\mathcal D|} &\sum_{(x,y_w,y_l)\in\mathcal D} \log \sum_{k=1}^K \big(\pi_{\phi,k}(x) \cdot\\ &\sigma\!(r_{\theta_k}(x,y_w) - r_{\theta_k}(x,y_l))\big). \end{aligned} \end{equation}

We adapt the interpretability regularization terms from~\citet{wang2026sparse} to encourage sparse, disentangled, and diverse \textbf{contribution patterns} in each expert.
Specifically, we define $\mathbf{c}(x,y_w,y_l)=[c_1,\ldots,c_K]$ as the vector to denote the absolute contribution contrasts of all experts.
When applying the local sparsity and global balance regularizations, we convert $\mathbf{c}(x,y_w,y_l)$ to a distribution $\tilde{\mathbf{c}}(x,y_w,y_l)$:
\begin{equation}
 \tilde{c}_k(x,y_w,y_l)
=
\frac{
c_k(x,y_w,y_l)
}{
\sum_{j=1}^{K} c_j(x,y_w,y_l) + \epsilon
}   
\end{equation}
and $\tilde{\mathbf{c}}(x,y_w,y_l)=[\tilde{c}_1,\ldots,\tilde{c}_K]$.
Then the CoCo-adapted interpretability regularization terms are defined based on $\mathbf{c}$ and $\tilde{\mathbf{c}}$. 

\noindent\textbf{(1) Local Sparsity}: 
We encourage sparse contribution contrast $\mathbf{c}(x,y_w,y_l)$ by minimizing the entropy of its normalized distribution $\tilde{\mathbf{c}}(x,y_w,y_l)$:
\begin{equation}
\begin{aligned}
\mathcal L_{\mathrm{ls}}
=
\mathbb{E}_{(x,y_w,y_l) \sim \mathcal D}
&\left[
\frac{H\!\big(\tilde{\mathbf{c}}(x, y_w, y_l)\big)}{\log K}
\right],
\\ 
H(\tilde{\mathbf{c}}) = -&\sum_{k=1}^K \tilde{c}_k \log \tilde{c}_k.
\end{aligned}
\label{eq:local entropy}
\end{equation}

\noindent\textbf{(2) Global Balance}: We encourage balanced utilization of experts to prevent model collapse. 
Specifically, we maximize the entropy of the average contribution contrast $\tilde{\mathbf{c}}(x,y_w,y_l)$ within a batch $\mathcal B$:
\begin{equation}
\begin{aligned}
\mathcal L_{\mathrm{gb}}
= 
\mathbb{E}_{\mathcal B \sim \mathcal D}
\left[
- \frac{H(\mathbb{E}_{(x, y_w, y_l) \sim \mathcal B}[\tilde{\mathbf{c}}(x, y_w, y_l)])}{\log K}
\right].
\end{aligned}
\end{equation}
\noindent\textbf{(3) Expert Diversity}: We also encourage experts to contribute diversely to the final reward by minimizing the average pairwise Pearson correlations between their $c_k$:
\begin{equation}
\mathcal L_{\mathrm{div}}
=
\frac{2}{K(K-1)}
\sum_{i<j}
\mathrm{corr}(c_i, c_j)^2,
\label{eq:expert correlation}
\end{equation}
where the Pearson correlations are computed over a batch $\mathcal B$ of preference pairs.

The final training objective is defined as:
\begin{equation}
\mathcal L
=
\mathcal L_{\mathrm{RM}}
+ \lambda_{\mathrm{ls}} \mathcal L_{\mathrm{ls}}
+ \lambda_{\mathrm{gb}} \mathcal L_{\mathrm{gb}}
+ \lambda_{\mathrm{div}} \mathcal L_{\mathrm{div}},
\end{equation}
where $\lambda_{\mathrm{ls}}$, $\lambda_{\mathrm{gb}}$, and $\lambda_{\mathrm{div}}$ control the strengths of local sparsity, global balance, and expert diversity regularization, respectively.

\end{document}